# Feature Detection for Hand Hygiene Stages

Rashmi Bakshi, Jane Courtney, Damon Berry, Graham Gavin

*TU Dublin*

**Abstract**

The process of hand washing involves complex hand movements. There are six principal sequential steps for washing hands as per the World Health Organisation (WHO) guidelines. In this work, a detailed description of an aluminium rig construction for creating a robust hand-washing dataset is discussed. The preliminary results with the help of image processing and computer vision algorithms for hand pose extraction and feature detection such as Harris detector, Shi-Tomasi and SIFT are demonstrated. The hand hygiene pose- Rub hands palm to palm was captured as an input image for running all the experiments. The future work will focus upon processing the video recordings of hand movements captured and applying deep-learning solutions for the classification of hand-hygiene stages.

# 1 Introduction

According to the European Centre for Disease Prevention and Control (ECDC), 2.5 million cases of Hospital Acquired Infections occur in European Union and European Economic Area (EU/EAA) each year, corresponding to 2.5 million DALYs (Disability Adjusted Life Year) which is a measure of the number of years lost due to ill health, disability or an early death [1]. MRSA-Methicillin Resistant Staphylococcus Aureus is common bacteria associated with the spread of Hospital Acquired Infections (HAIs) [2]

One method to prevent the cross transmission of these microorganisms and to reduce the spread of HAIs is the implementation of well-structured hand hygiene practices. The World Health Organization (WHO) has provided guidelines about hand washing procedures for health care workers [3]. Best hand hygiene practices have proven to reduce the rate of MRSA infections in a health care setting [4].

One potential approach is to use camera system or 3D gesture trackers to track fine hand movements and identify user gestures, provide feedback to the user or a central management system, with the overall goal being an automated tool that can ensure compliance with the hand washing guidelines. In advance of developing these systems, however, preliminary analysis for hand features extraction and detection was required.

The aim of this paper is to extract contours, apply popular feature descriptors such as Harris, Shi-Tomasi, and SIFT features to hand hygiene images. A robust dataset of video recordings is prepared for the classification of hand hygiene stages as a future work based on the paradigms of Deep Learning.

# 2 Feature Detection

A feature is an attribute or a characteristic property of an object, which is tracked. Corners, Edges and blobs are three types of image features in the field of computer vision [6].

A corner is an intersection of the two edges; it represents a point in which the directions of these two edges change. In other words, a corner represents the variation of intensity in an image.

## 2.1 Corners vs Edges

Corners are the locations where variations of intensity function f(x, y) in both X and Y are high. Both partial derivatives fx and fy are large. Edges are the locations where variation of f(x, y) in certain direction is high, while variation in the orthogonal direction is low. If an edge is oriented along Y, fx is large and fy is small. Figure 1 differentiates between a flat region, an edge and a corner by demonstrating a change in the directions. For instance, a flat region does not change in all directions. An edge are those pixel values that do not change along an edge direction. A corner are those points that have significant variation in all the directions and therefore suitable for image classification [6]

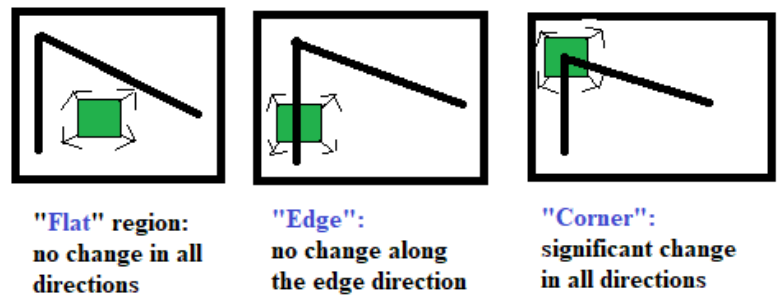

**Figure 1: A flat region, an edge and a corner**

## *Properties of a strong feature*

To build a robust hand feature detector based on camera images, one needs to understand and differentiate strong features from weak features. The general properties of a strong feature are:

- Features are local and accurate
- Features are robust (to noise, blur, compression)
- Distinct: Individual features can be matched to a large database.
- Invariant: where a feature does not change in case of rotation, image scaling and lighting difference. For instance, a 'corner' identified as an 'edge' if the scale of an image changes is an example of a poor feature [6].

Corners are regarded as "good features" in comparison to edges as they are uniquely identifiable due to large variations in all directions.

There are various feature detection algorithms in the field of computer vision. In this paper, Harris feature detector, Shi-Tomasi and SIFT detector were applied for extracting hand features in a hand pose-*Rub hands palm to palm*. A keypoint extraction algorithm, Harris Corner detector [5] used the combination of corners and edge points to describe the features but it fails to describe surfaces or an object as a whole. Shi-Tomasi is based on Harris detector with different scoring criteria for identifying the interest points and proves to perform better than Harris detector [6]. Scale Invariant Feature Transform (SIFT) [9] algorithm was introduced in 2004, a huge paradigm shift in feature extraction and description. SIFT addressed the challenge of invariance and it used difference of gaussian function for detecting potential keypoints and used image gradient magnitude, direction from local neighbourhood for keypoint description. It is widely used for object detection and image matching tasks but considered as computationally expensive and therefore unsuitable for real-time applications [8]. Other widely known feature descriptors are Speeded up robust features (SURF), Features from accelerated segment test (FAST) and Binary robust elementary features (BRIEF) and a combination of Oriented FAST and rotated BRIEF (ORB) [7,8] . SURF outperforms all the above features and is known as computationally efficient [8].

# 3 A Rig Construction for Dataset Preparation

A fixture rig was built; a prototype is shown in Figure 2, with the purpose of mounting cameras and gesture tracker near a sink. It was used for recording different hand movements, hand-washing stages and in gathering videos for a later dataset.

An aluminium rig was built with the dimensions of 1x0.8x0.8 m (LxWxH). The measurements were selected to ensure enough space to fit on a sink and accommodate utilities such as soap dispenser, hand-towel.

A controlled background exposure was selected to avoid the presence of skin coloured objects that can be miss-classified as an actual skin. Green and white sheets were used to minimise the unnecessary background information during the data collection procedure. The participants were anonymous and only hand movements were captured. The height of the frame was reduced from 1 m to 0.8 m in order to avoid the appearance of body organs in the frame other than the hands.

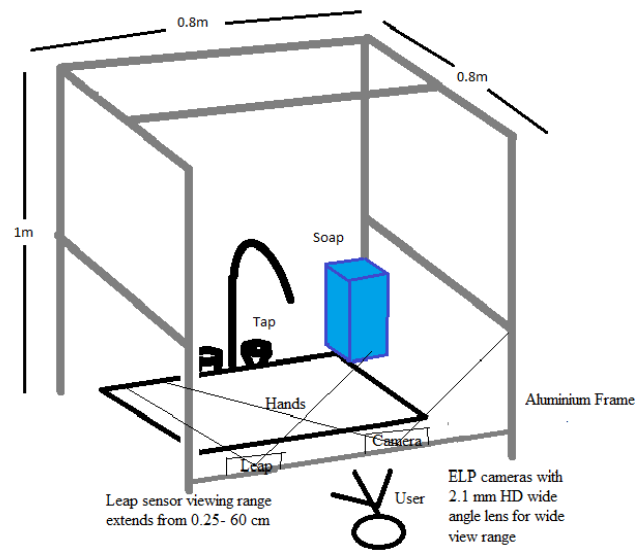

**Figure 2: An aluminium rig for data collection**

## 3.1 Hand-washing dataset

The importance of having a video data set or repository was identified and 30 volunteers were asked to perform the various hand washing gestures in this project. In addition to the hand washing movements, one-hand gestures such as linear and circular hand movements were also recorded. The aluminium rig was placed over the laboratory's sink and a digital camera was mounted over it. The camera was mounted in such a way that only the hand movements of the participants were recorded. This was done to maintain the anonymity of the user. The participants were informed about the study and were asked to complete an information sheet. Before recording the videos, the participants were informed about the importance of hand washing in a healthcare setup; WHO hand hygiene stages were demonstrated. They were able to refer to the hand hygiene poster while performing the hand washing movements. However, the participants were not enforced or trained to allow them to carry out the hand washing steps in a natural manner with their personalised interpretation of the guidelines. The video length for the hand washing activity was recorded for 25-30 seconds. Every hand-washing step was followed by a pause where in the user was instructed to move their hands away from the camera. Video format for this data set is MP4 file with a size of range 40-60 MB and a frame rate of 29.84 frames/s. All of the six hand washing movements were recorded in one video for each participant.

An example of the collected information in addition to the video recording, stored in a csv file for all the participants

```
Gender = Male, Age = 29, Profession = Researcher, Country of origin =
India, Skin tone = Brown, Video size = 62.9 MB, Video length = 29 seconds
```

## 4 Methodology and Results

Python based OpenCv library is used for conducting the experiments. Hand-hygiene pose- Rub hands Palm to Palm was taken as an input image in order to run the python scripts for the algorithms described below.

Hand hygiene video recording-Rub hands palm to palm was captured and utilised to extract centroid information. The resultant image is shown in Figure 6.

Algorithm-1 Contour, Convex Hull Detection

```
Read input image in RGB format
Convert the image to Gray-scale format
Apply Gaussian blur to reduce the extra noise
Threshold the image
Find contours in the image
Sort the contours and find the largest contour
Find convex hull for the largest contour
Display the image with contour and convex hull
```

Algorithm-2 Harris Detector

```
Read input image in RGB format
Convert the image to Gray-scale format
Apply cv2.cornerHarris function.
Dilate the image for marking the corners
Threshold for an optimal value
Save the image
```

Algorithm 3- Shi-Tomasi Detector

```
Read the image
Convert the image to Gray-scale format
Detect the corners by applying goodFeaturesToTrack function
Mark the corners with circles
Save the image
```

Algorithm 4- SIFT Keypoints Detector

```
Read the input image
Convert the image to Gray-scale format
Apply xfeatures2d.SIFT_create()
Detect Keypoints in an image
```

```
Display the Keypoints

Save the image
```

| Feature Detector | Rotation invariance | Scale invariance | Illumination invariance |
|---|---|---|---|
| Harris | Yes | No | No |
| Shi-Tomasi | Yes | No | No |
| SIFT | Yes | Yes | Yes |

**Table 1**

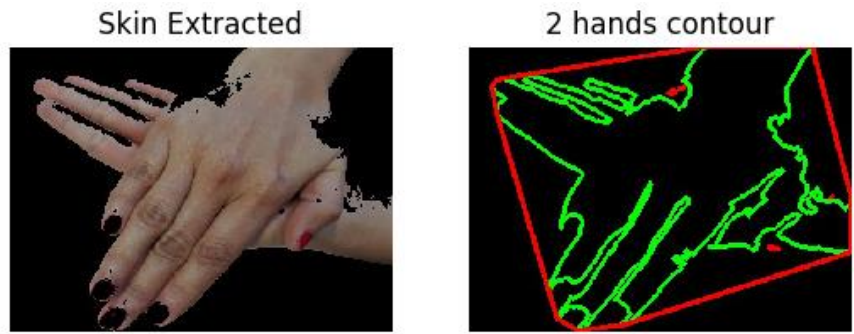

**Figure 3: Skin extracted for Rub Hands Palm to Palm (L); Contours-convex hull (R)**

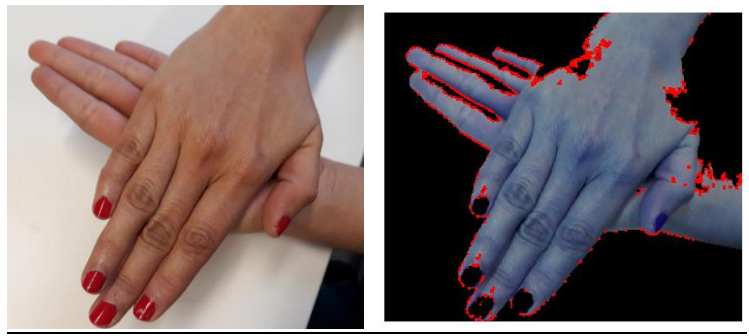

**Figure 4: Rub Hands Palm to Palm (L); Harris features detected (R)**

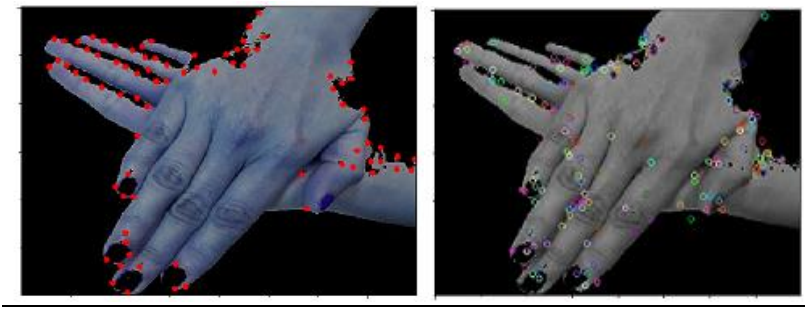

**Figure 5: Shi-Tomasi features (L) and SIFT features (R)**

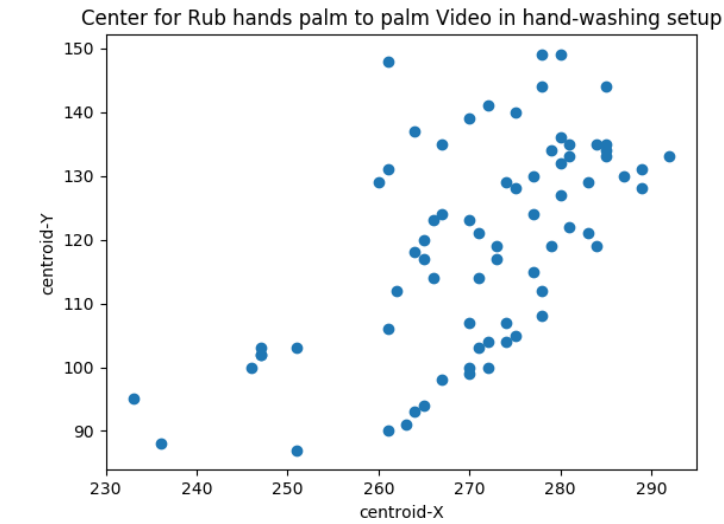

**Figure 6: Centroid(x, y) extracted for hand hygiene video recording-Rub hands palm to palm**

## Future work

The hand hygiene video repository was created by recording hand movements of 30 participants in a controlled environment. The video recording comprise of six hand hygiene stages clearly separated from one other. The videos can be decomposed to extract each stage, assign a label and process them individually.

With advancement in computational resources and availability of the data, deep learning techniques are becoming popular for object detection, recognition and other computer vision tasks. Deep neural network models like Inception, VGG, XCeption, ResNet [10] are already being implemented and tested on a variety of datasets. With the use of transfer learning, tune the fine parameters of these pre-trained models, various small datasets are utilised and tasks completed. The future work will focus upon applying the deep learning route- training the pre-trained CNN model with the video/ image dataset with distinct number of classes and making the predictions on the unlabelled data that the network has not seen before. The loss-accuracy curve will be plotted for the trained and validation data in order to ensure that the network is trained correctly.

7. F. K. Noble, "Comparison of OpenCV's Feature Detectors and Feature Matchers." IEEE 23rd International Conference on Mechatronics and Machine Vision in practice, China, 2016.
8. J. Borse, D. Patil, and V. Kumar, "Tracking Keypoints from Consecutive Video Frames Using CNN Features for Space Applications Tracking Keypoints from Consecutive Video Frames Using CNN Features for Space Applications," Technik, glasnik, Vol. 15, No. 1, 2021.
9. Lowe, D. G. (2004). Distinctive Image Features from Scale-Invariant Keypoints. *International Journal of Computer Vision, 60*, 91-110.
10. Keras Applications. Available models. https://keras.io/api/applications/#available-models Accessed: 2021